\documentclass{article}

\usepackage{microtype}
\usepackage{graphicx}
\usepackage{subfigure}
\usepackage{booktabs} % for professional tables

\usepackage{multirow}
\usepackage{amsmath}

\usepackage{adjustbox}
\usepackage{hyperref}

\usepackage[accepted]{icml2019}

\icmltitlerunning{Semi-Supervised Class Discovery}
\begin{document}

\twocolumn[
\icmltitle{Semi-Supervised Class Discovery}



\icmlsetsymbol{equal}{*}

\begin{icmlauthorlist}
\icmlauthor{Jeremy Nixon}{equal,goo}
\icmlauthor{Jeremiah Liu}{gor}
\icmlauthor{David Berthelot}{goo}
\end{icmlauthorlist}

\icmlaffiliation{goo}{Google Brain, Mountain View, California}
\icmlaffiliation{gor}{Google Research, Mountain View, California}

\icmlcorrespondingauthor{Jeremy Nixon}{jeremynixon@google.com}
% \icmlcorrespondingauthor{Eee Pppp}{ep@eden.co.uk}

\icmlkeywords{Machine Learning, Semi-Supervised Learning, Clustering}

\vskip 0.3in
]



\printAffiliationsAndNotice{}  % leave blank if no need to mention equal contribution

\begin{abstract}
One promising approach to dealing with datapoints that are outside of the initial training distribution (OOD) is to create new classes that capture similarities in the datapoints previously rejected as uncategorizable. Systems that generate labels can be deployed against an arbitrary amount of data, discovering classification schemes that through training create a higher quality representation of data. We introduce the Dataset Reconstruction Accuracy, a new and important measure of the effectiveness of a model's ability to create labels. We introduce benchmarks against this Dataset Reconstruction metric. We apply a new heuristic, class learnability, for deciding whether a class is worthy of addition to the training dataset. We show that our class discovery system can be successfully applied to vision and language, and we demonstrate the value of semi-supervised learning in automatically discovering novel classes.
\end{abstract}

\section{Introduction}
% Something boring about how labeling is expensive
% Something cool about the potential of a feedback loop between representation learner and data
% Add content that's the intersection of introductions from related papers

An ongoing transition in machine learning research has been from hand-labeling datasets from which our models learn to having an algorithm create a dataset from which our models learn. The process of dataset creation is time consuming, expensive and ad-hoc. Progress in self-supervised \citep{radford2019language} \citep{devlin2018bert} and semi-supervised \citep{Berthelot_undated-jm} learning has enabled us to pre-train in an infinite data training regime where the main limit to our training is computation time. They've led us to techniques that allow us to automatically label unlabeled data points. In contrast with the standard semi-supervised approach of taking advantage of additional samples whose classes are already known, we take advantage of data that is from classes that have never been seen at a model's original training time, creating a learning algorithm that automatically generates an effective learning environment.

\begin{figure}[t]
\includegraphics[width=.99\columnwidth]{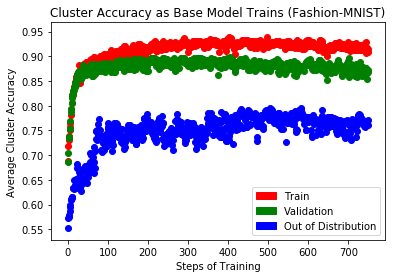}
\includegraphics[width=.49\columnwidth]{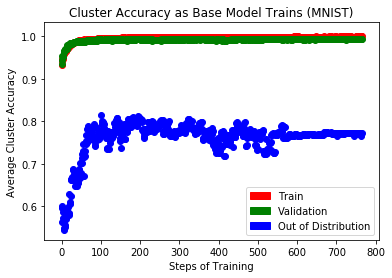}
\includegraphics[width=.49\columnwidth]{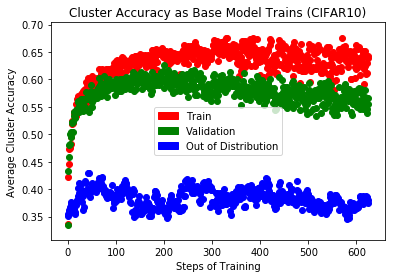}
\caption{Over the course of training on supervised labels, the ability to discover quality clusters in out-of-distribution data points improves. The red and green curves measure the correspondence of in-distribution clusters to the original class that has the maximum overlap with the cluster. The blue curve measures the correspondence of out-of-distribution clusters to the held-out classes.}
\end{figure}

One major advantage of this technique is transforming out-of-distribution data into in-distribution data. Data that is out-of-distribution leads to an inability to make claims about a model's confidence or performance on that data, making' it hard to trust your model's confidence and calibration \citep{snoek2019can}. Bringing out-of-distribution data points into the training dataset and training on them is one way to rectify uncertainty estimates in a multi-class setting where new classes can be productively added to the training distribution.

\begin{figure}[t]
\includegraphics[width=.99\columnwidth]{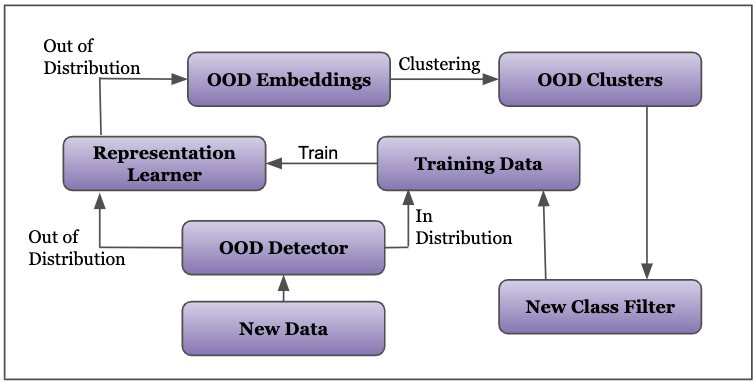}
\caption{Diagram of this dynamic self-supervised learning system. Training data is fed into the representation learner. The representation learner trains on the data, improving its performance on its present labeled tasks. New data is processed by an OOD detector that classifies the data as in-distribution (in which case it enters the training data in its corresponding class) or out-of-distribution. If the data is out-of-distribution, it is fed through the representation learner and embedded through to the penultimate layer (the activations before data is fed to the dense layer that leads to the softmax). Those embeddings are clustered (in our case by k-means, but it could be any clustering algorithm). Each cluster is evaluated by training a classifier (a small CNN) to predict the cluster for each datapoint. The data from the cluster that is most easily learned and for which the classifier gets the highest accuracy is accepted as a new class and is added to the training data, along with a new and additional label. The model then continues to train on all data, including the new data which used to be out-of-distribution and which now is in the training distribution.}
\end{figure}

The self-supervised infinite training data regime has been a boon to transfer learning research \citep{raffel2019exploring}. The features developed in training on classes similar to the classes to be discovered aid in accomplishing faster learning on those classes. We accomplish a kind of transfer learning in line with self-taught learning \citep{raina2007self}, where data from classes not found in training improve model performance. Much of the success of deep learning in vision can be attributed to the availability of large-scale labeled data \citep{sun2017revisiting}, and so developing the ability to construct large-scale datasets is incredibly important.

This approach can be seen as conditional generative modeling for labels, where we condition on the existing labeled data. This allows the trained model to learn a prior for data representation that comes from human labels, where the trained model pushes classes with human labels to be linearly separable in a way that can generalize to data that hasn't been labeled, imitating the heuristics that a human uses to select and distinguish between classes. In contrast, unsupervised data labeling techniques can't impose a prior which is influenced by what a human would have done. One major difference from generative modeling is that here, we generate partitions. One partition is over the in-distribution space, covered by existing classes and their datapoints. The other partition is over the out-of-distribution space, which is a partition generated by a clustering algorithm. That partition will progressively merge with the in-distribution partition.

There is evidence that dataset size is still a major limiting factor to the effectiveness of deep learning models \citep{sun2017revisiting}.
One reason much work in vision is on the classification task (as opposed to the more general task of identifying all objects in an image) is that it's challenging to get quality labeled datasets. Techniques like class discovery can help overcome that challenge.

Our major contributions are demonstrating the value \& potential of semi-supervised learning backed by deep metric learning in class discovery, demonstrating a method that is general across language and vision, introducing a new class feature which can be used to distinguish class quality, introducing the Dataset Reconstruction Accuracy measure, and demonstrating the value of creating a feedback loop between learning on discovered classes to create a better model for representing OOD data and using that improved representation to discover classes more effectively.

\section{Related Work}

% [TODO: jeremynixon] Make this section into prose, and cut out the unnecessary cites.

\subsection{Semi-Supervised Learning}
There is a recent wave in approaches for self-supervised and semi-supervised learning \citep{Berthelot_undated-jm}, which are relevant because our method is initialized in a semi-supervised fashion before it continues in a self-supervised fashion. Our method, however, differs dramatically from regular self-supervised learning where all classes are present in the training dataset and the model merely learns to slot unlabeled datapoints into those pre-existing classes.

In regular semi-supervised learning, the data set $X = (x_i)_i \in[n]$ can be divided into two parts: the points $X_l := (x_1, ..., x_l)$ for which labels $Y_l := (y_1, ..., y_l)$ are provided, and the points $X_u := (x_{l+1}...,x_{l+u})$ that are in the same domain as $X_l$ but the labels of which are not known.

In contrast with conventional semi-supervised learning, we consider $X_{u}\in \textit{OOD} := (x_{l+1} \in \textit{OOD}, ..., x_{l+u} \in \textit{OOD})$ for which the data is not necessarily in the same domain as the $X_l$, and a correct existing class label is not available to the model.

\subsection{Deep Metric Learning}
We employ deep metric learning, using data to learn a similarity measure on which we cluster. Clustering in deep networks has occurred in other contexts. 
Learning Discrete Representations via Information Maximizing Self-Augmented Training (IMSAT) uses deep clustering to learn a discrete representation that is invariant to perturbations \citep{Hu2017-bh}. Deep Clustering for Unsupervised Learning of Visual Features improves representation quality by creating labels for existing data points and training on said labels, though they do not train on datapoints outside the original training distribution. Their work offered initial proof that representations can be improved by training on classes discovered via clustering \citep{Caron2018-am}.

Deep Metric Learning via Facility Location \citep{Song2016-fc} introduced Normalized Mutual Information, a measure for comparing discovered classes to existing class labels.

% Generalized Clustering by Learning to Optimize Normalized Cuts (not yet published)\\\\
% Local Aggregation for Unsupervised Learning of Visual Embeddings\\
% Visual Similarity from Optimizing Feature and Memory On A Hypersphere \\
% Stacked Capsule Autoencoders\\
% Invariant Information Clustering for Unsupervised Image Classification and Segmentation\\

\subsection{Open World Classification}
Our system is related to the closed-world assumption \citep{scheirer2012toward}, where our classifier does not make the assumption that all classes which appear in the test data must have appeared during training. Our system stretches this to a case where the initial training distribution does not include all classes, but where classes are introduced to the training distribution after being created by our model. Unseen Class Discovery in Open World Classification unmakes this assumption but focuses on finding the correct number of outstanding classes, rather than finding the data points which belong to those classes. \citep{Shu2018-lv} Towards Open World Recognition \citep{Bendale_undated-by} sets the frame for engineering open world systems but does not discover new object categories itself, leaving that to human labeling. iCaRL: Incremental Classification and Representation Learning \citep{Rebuffi_undated-jv} focuses on incremental concept discovery, but does not discover the classes itself. Simultaneous Class Discovery and Classification of Microarray Data Using Spectral Analysis \citep{Qiu2009-bx} both discovers classes (with a spectral analysis technique) and classifies those classes. No deep metric learning is used to represent the data, but the application to microarray data is an example of the value of the technique.

\subsection{Bayesian Nonparametrics}
The problem of generating new labels has been touched on in Bayesian Non-parametrics, including Distance Dependent Chinese Restaurant Process \citep{Blei2011-bb} and Latent Dirichlet Allocation (LDA) \citep{Blei2003-qp}. We compare against LDA while applying our method to the Clnic OOS dataset.

\subsection{Continual and Lifelong Learning}
We present a continual learning and lifelong learning system, which is influenced by the continual learning frontier. For example, Continual Unsupervised Representation Learning \citep{rao2019continual} aims to create an algorithm that discovers new concepts over its lifetime, using a generative model of past classes to avoid catastrophic forgetting. We show the advantage of leveraging a combination of human-labeled data and algorithm-labeled data in a similar setting.
% Don’t forget, there is more than forgetting: new metrics for Continual Learning\\
% Measuring Cumulative Gain of Knowledgeable Lifelong Learners\\
% Continual Learning with Self-Organizing Maps\\
% ContinuLearning in Practice\\
% Continual Unsupervised Representation Learning (Evaluates Cluster Accuracy [code])\\
% Continual Learning Benchmarks\\

\subsection{Self-Supervised Learning}

Self-Supervised learning has seen lots of recent success, especially in language \citep{radford2019language} \citep{devlin2018bert}. Extensions to vision have mainly focused on pretext tasks like prediction context \citep{doersch2015unsupervised}, image rotation \citep{idaris2018unsupervised}, or architecture design \citep{kolesnikov2019revisiting}.

Our work is inspired by the transition to the pre-training with infinite training data regime achieved by these models, though it creates a task by using a system of machine learning algorithms (embedding + clustering) to generate the label rather than taking it directly from the training data. This differentiates of from other mergers of self-supervised and semi-supervised learning, such as Self-Supervised Semi-Supervised Learning \citep{Zhai2019-kd}.

% An Overview of Proxy-Label Approaches for Semi-Supervised Learning [Ruder]\\
% Strong Baselines for Neural Semi-supervised Learning under Domain Shift [Ruder]\\
% Semi-Supervised Learning Textbook [Compilation]\\
% Semi-Supervised Clustering\\
% Semi-Supervised Textbook Chapter on Probabilistic Semi-Supervised Clustering with Constraints\\
% Cases where complete class labels may be unknown\\
% Learning a Mahalanobis Metric from Equivariance Constraints\\
% Speaker identification, as well as finding a good distance metric.\\
% Constrined K-means Clustering with Background Knowledge\\
% Realistic Evaluation of Deep Semi-Supervised Learning Algorithms\\
% Classic Semi-Supervised Task Papers\\
% Mix-Match\\
% Revisiting Self-Supervised Visual Representation Learning\\
% Self-Supervised Semi-Supervised Learning\\
% Data-Efficient Image Classification with Contrastive Predictive Coding\\
% GPT2\\
% BERT (inspiration for automatically generating training data)\\
% Multi-task Self-Supervised Visual Learning
% Unsupervised Learning by Predicting Noise\\
% Learning Representations by Maximizing Mutual Information Across Views\\

\subsection{Semi-Supervised Clustering}

Metric-based semi-supervised clustering also focused on using some labeled data to aid unsupervised learning. There was rich activity in the field in the early 2000s (Ex., \citep{bilenko2004integrating}). Rather than learning new classes, the goal is to fit existing classes well and learn new clusters around them. Metrics were often learned with an SVM and deep metric learning was rare.

\subsection{Task Generation}
There is a thread in intrinsic motivation research where an agent is trained to generate new tasks that it is capable of performing in an environment and then learn on those tasks \citep{held2018automatic}, \citep{schmidhuber1991curious}, \citep{schmidhuber2013powerplay}, \citep{clune2019ai}. Our system has similar properties and motivations, though we focus on the multi-class classification task. 

\section{Method}

 As depicted in Figure 2, our method consists of a representation learning, an out-of-distribution (OOD) detector, generated OOD embeddings, a clustering algorithm generating clusters of OOD embeddings, a filtering of those clusters as new class candidates, and a dynamically updating training dataset from which the representation learner learns. We describe many of those parts in detail in the following sections.

\subsection{Representation Learner}
Our representation learner is trained to classify over all of the current training data. It will create embeddings of our OOD data, whether the domain is images or language. The  representation learners employed include a small CNN and Resnet50 \citep{he2016deep} for image data, and the TextCNN \citep{kim_convolutional_2014} for text data.

\subsection{OOD Detector}
We take advantage of \citep{hendrycks2016baseline}'s work in out of distribution detection, which evaluates a given data points as being whether out-of-distribution by thresholding its predictive confidence (i.e., the maximum predictive probability) with a pre-computed cut-off value. Specifically, this method computes the cut-off value as the 95\% quantiles of predictive confidence of the training data, and using that confidence level to evaluate new data as being out of distribution. For example, our model may show that 95\% of training datapoints have a class that is predicted with a probability of 80\% or above. We would then treat new datapoints whose top class is predicted with 80\% or less as out of distribution. In general, progress in OOD Detection (Ex. \citep{Liang2017-nu, Lee2018-cv}) will lead to more effective class discovery systems, as more data points will correctly be placed into the data store from which new classes are created.

\subsection{Discrete Class Creation}
We use the clustering method K-means to transform a high dimensional set of embeddings into new classes to be evaluated. K-means gives a higher quality of candidate classes, as measured by correspondence to the original class set in comparison to other clustering methods like Hierarchical Agglomerative Clustering. Hierarchical clustering returns a hierarchy with multiple levels of class candidates, which allows for mode discovery and a choice of what level at which to define a label set. For us, this was not worth the drop in the quality of discovered clusters.

We initialize k-means with a k-means++ initialization, focusing the initial clusters on regions that are far from one another.

We run k-means 10 times with different centroid seeds, using the result that performs best on a measure of inertia. Inertia is the sum of squared distances of samples to their closest cluster center.

\subsection{Class Addition Heuristic}

% \begin{figure}
% \includegraphics[width=.99\columnwidth]{figures/class_accuracy.png}
% \caption{A surprising negative result: One would expect a cluster's density to be a strong predictor of that cluster's correspondence to the unseen true class labels, as the density should indicate similarity between the elements in the cluster. Empirically, cluster density, is a poor predictor of cluster accuracy.}
% \end{figure}

Carefully choosing which new class candidates are worth adding to the training dataset is important to preserving the quality of the training set. We explore three major heuristics. The first is class learnability, which asks whether a model trained on the candidate classes can effectively learn to predict whether a datapoint is from the candidate class. Here we ask if a class, when added to the training dataset, leads to an improvement in the ability of the representation learner trained on that dataset to quickly adapt to new tasks. The third is a measure of cluster density, assuming that denser clusters are more likely to contain datapoints that are similar to each other and so will have a higher quality. This led to a negative result, where denser clusters were not related to cluster accuracy. One important realization is that all three are features of clusters. The can be used in tandem or be put through a machine learning model that predicts a cluster's value.

\subsection{Evaluating Performance with Dataset Reconstruction Accuracy}

We use two major measures to evaluate class discovery performance in this paper.

 The first major evaluation measure is cluster accuracy. We measure the amount of overlap that a given out-of-distribution cluster has with the removed class that it has the most overlap with. For example, an MNIST class with 6 '9' values and 4 '7' values would be mapped to class 9 and given a cluster accuracy of 60\% (6/10). Multiple clusters are allowed to be mapped to the same unseen original class label.

Second, we introduce the dataset label reconstruction accuracy to measure the efficacy of our class discovery  system. Dataset reconstruction asks what the overall overlap of the original class labels are with the labels discovered and assigned during class discovery. This overlap score is determined by maximum overlap. For example, if a cluster is discovered, the entire cluster is assigned to a new label. The check against the original dataset comes from comparing with the ground truth labels of the datapoints for each class. The class with the plurality of labels represented has every datapoint in the new cluster assigned to it. 

One very important advantage of the dataset reconstruction score over cluster accuracy is its generality. It's easy to imagine class discovery systems being created, for example, via raw discrete latent variable models. In that case, comparing discrete latent variable models to this technique would require a measure that wasn't centered on clustering.

Let training labels $Y_\ell := (y_1, ..., y_\ell)$ where $\ell$ is equal to the number of training datapoints. Let OOD labels $Y_o := (y_t,..., y_t+o)$ where $o$ is equal to the number of OOD data points whose correct class label is not available to the model. and which correspond to the labels that the system discovered itself. Let the normalized cluster size $\boldsymbol{w}$ be the weighting of each discovered cluster based on the number of datapoints in the cluster and $\boldsymbol{a}$ be the accuracy of those clusters, which is the maximum overlap against the true label set. We compute the dataset reconstruction accuracy through an indicator function for whether a discovered label matches the true label distribution or not, which is

$$ \frac{1}{N}\sum_{i=1}^N I(y_{\text{discovered}, i} = y_{\text{true}, i}) $$

In our case, we use clustering to compute the datset reconstruction accuracy over K clusters, giving us:

$$ \frac{\ell + o * (\sum_{k=1}^K w_k*a_k)}{N}  $$

\section{Experiments}

\begin{table}
    \centering
    \textbf{Cluster Accuracy vs. Class Count during Training} \label{tab:title} 
    \begin{tabular}{|c|c|c|c|}
        \hline\hline
        Class Count & MNIST & Fashion-MNIST & CIFAR10 \\ \hline
        2 & 0.4282 & 0.5063 & 0.3536  \\ \hline
        3 & 0.4958 & 0.6495 & 0.3403 \\\hline
        4 & 0.6497 & 0.6931 & 0.3971\\ \hline
        5 & 0.7197 & 0.7005 & 0.4013 \\
     \hline 
    \end{tabular}
    \caption{Training a small CNN on more initial classes leads to stronger OOD Cluster Accuracy scores. This is  after controlling for datapoint count and is always evaluated on 5 outstanding classes.}
    \label{tab:class_count}
\end{table}

\subsection{Datasets}
We evaluate the method on both image and language modalities. We consider MNIST \citep{lecun1998mnist}, Fashion-MNIST \citep{xiao2017fashion}, CIFAR 10 \citep{krizhevsky2010convolutional} for image, and CLINC out-of-scope (OOS) intent detection benchmark \citep{larson_evaluation_2019} for language.

There are a few settings for experiments, where our datasets are split according to the experiment. In the default setting, half of the classes of (MNIST, Fashion-MNIST, CFAIR10) have the labels stripped from their datapoints. In self-supervised style, a supervised training set is composed of the other 5 classes. An oracle correctly determines the split between in distribution data and out-of-distribution data.

In the second setting, an OOD detector determines whether data points are slotted into existing classes or pooled with other out of distribution data points for assignment to new class labels. This is required for the dynamic model because if it creates two classes which are have an identical ground truth label it will struggle to differentiate them from one another and training will stall.

On the CLINC OOS dataset, 120 classes are used during training and 30 classes are used for class discovery evaluation, as well as an out-of-scope class which contains random user utterances that do not match any of the known intents.

\subsection{Number of Initial Classes Effects}
This important experiment shows that if you are able to successfully label new classes, your ability to discover new classes improves. The difference between training on 2 classes and training on 5 classes led to an increase from 43\% accuracy to 71\% accuracy in the small CNN case on MNIST (See Table 1). This dramatic impact from the addition of training data shows the potential of effective class discovery systems that dynamically feed back on themselves as they become more capable of discovering classes with each new class they discover.

\begin{table}
    \centering
    \textbf{Dataset Reconstruction Accuracy during Discovery}
    \begin{tabular}{|c|c|c|c|}
        \hline \hline
        Classes Discovered & MNIST & Fas.-MNIST & CIFAR10 \\ \hline
        No Added Classes & 0.8473 & 0.8290 & 0.6714  \\ \hline
        Added 1 Class & 0.8813 & 0.8549 & 0.6796 \\\hline
        Added 2 Classes & 0.8958 & 0.8578 & 0.6921\\ \hline
        Added 3 Classes & 0.9054 & 0.8589 & 0.7053 \\
        \hline
        Added 4 Class & 0.9078 & 0.8644 & 0.7188 \\ \hline
        Added 5 Classes & 0.9113 & 0.8695 & 0.7319 \\ \hline
     \hline
    \end{tabular}
    \caption{Improvement in data set reconstruction accuracy through training on newly discovered classes. Note that these scores are the total fraction of correctly recovered labels in the training dataset (modulo ordering), and not the average correspondence of cluster labels to the original OOD set.}
    \label{tab:discovery}
\end{table}

\subsection{Other Experimental Methodology}

The representation learner used in our experiments is either Resnet50 \citep{he2016deep} or a small CNN. The small CNN has a single convolutional layer with 32 filters and a Relu activation, followed by a flattening and a dense layer with 128 neurons and a relu activation.

We apply an Adam optimizer with a learning rate of $0.001$, beta 1 of $0.9$, beta 2 of $0.999$, and epsilon of $1e-0.7$. Our error metric is the cross entropy loss function. 

Our OOD baseline detector has its threshold set to 95\% confidence on the training data.

The number of clusters $k$ in all vision experiments is 15. This hyperparameter has a fairly strong effect on both cluster accuracy (higher accuracy with higher $k$) and on dataset reconstruction score, so it's kept constant through baseline comparisons.

We use Tensorflow 2.0 for all of our neural network models \citep{abadi2016tensorflow}. For our linear models and clustering we use Scikit-Learn \citep{pedregosa2011scikit}.

An optimial hyperparameter sweep (ex., for the number of clusters or hyperparameters for the representation learner) would optimize for the dataset reconstruction accuracy in the setting where you're limited to creating a number of classes that is equal to the number of classes removed from the dataset.

All models were trained on a single GPU or CPU at a time. Our main compute infrastructure was a set of Nvidia p100 GPUs.

\subsection{Compared Methods}

We compare 3 methods for generating new classes via clustering, assuming an oracle allows you to correctly determine which datapoints are not from the existing class label set. These results are displayed in Table 3.

% 'Raw clustering' runs K-means in the pixel space over a flattenened representation of each image. 
'Random embedding' uses an untrained single convolutional layer followed by a single feedforward layer to embed the image. 'Semi-Supervised' trains that same model on the first 5 classes from each dataset. 'Dynamic' trains on the first 5 classes for an epoch, discovers and adds the most learnable cluster, and then trains on the entire dataset until 5 new classes have been added. 
% 'Radial Basis' uses a radial basis layer stacked on the CNN to embed the data.

\subsection{Static vs. Dynamic Class Discovery}

One simple approach to class discovery is to create classes over all the out of distribution datapoints at once, assigning every OOD datapoint to a new label. That approach can be made dynamic by choosing (for example) one new class at a time. Once that class is chosen, a re-training process integrates that class into the existing representation learner. That retrained model is used to generate and select another class, in an ongoing process.

We show results from both setups. In the static setting, an overall dataset label reconstruction accuracy on MNIST of 84.73\%. In the dynamic setting, we see an overall dataset label reconstruction accuracy of 91.13\%.

As the dynamic model adds and trains on new classes, its ability to recover the labels over the entire training dataset improves.

\subsection{Learnability as a Class Evaluation Technique}

% \begin{figure}
% \includegraphics[width=.99\columnwidth]{figures/log_reg_learnability.png}
% \caption{An application of logistic regression to predict clusters that are candidates for transformation into classes. Clusters that are learnable (which the model's accuracy is high for) are more likely to have a high correspondence with the true label set.}
% \end{figure}

One major challenge is finding new classes that are pure, where most data points correspond to a coherent concept. If discovered classes are impure, the training process will encounter incorrect and noisy labels. Training on those labels can be difficult, and lead to a degradation in training accuracy as well as a slower training process \citep{zhang2016understanding}. Do note, however, that the incorrect labels will be closer to correct than randombly permuted labels, because they will have been close datapoints in the embedding space.

We introduce a new technique, learnability, as a test of the quality of a cluster which could be added as a new class. The heuristic is as follows: Treat all of the clusters that come out of the OOD dataset as classes to be learned by a fresh, untrained model. Train that model on the clusters, treating the clusters that can be easily learned (for which you obtain a high accuracy) as better options for inclusion that clusters that are challenging to learn.

This will lead to selecting new classes that are separable from the rest of the data.
Including existing classes in this set is also an option, which can check for the separability of discovered classes with existing classes as well as separation from potential classes.

Using learnability, MNIST reconstruction accuracy is $0.9113$. Choosing random clusters leads to an accuracy of $0.9004$. Fashion-MNIST learnability reconstruction accuracy is $0.8695$, while the random baseline gives $0.8537$. The difference on CIFAR 10 is $0.7319$ to $0.6841$.

% \subsection{Isotropic Layer for Embedding Quality}

% % \begin{figure}
% % \includegraphics[width=.99\columnwidth]{figures/isotropic_fmnist.png}
% % \caption{Discovering classes using our Isotropic Layer leads to a substantial improvement in the discovered clusters' overlap with the hidden true class distribution.}
% % \end{figure}

% We apply a radial basis function transformation to make the embedding amenable to clustering which implements the function:

% $$ -\beta \frac{1}{N} \sum_{i=1}^N (\boldsymbol{x}-\boldsymbol{K})^2 $$

% Where $\beta$ represents the bias and K represents a kernel with the same shape as the pre-softmax layer.

% Embeddings are widely used for a variety of tasks: analogy reasoning, clustering, classification from pre-learned embeddings. However, due to the process through which embeddings are learnt, they may not be well suited by design for such tasks. An experiment from qincao@ demonstrated that taking embeddings on a learned ImageNet model and knowing the perfect assignment by looking at the class labels, optimal clustering of the embeddings resulted in much worse accuracy than classification.
% Can embeddings be improved for clustering and what are the effects on other tasks?

% Yes, embeddings can be improved and designed to work well for clustering. One solution is to make the embedding space isotropic, which can be solved with a new differentiable layer (isotropy layer). This has been verified on cifar10 (dberth@), learning an isotropy layer allows to scale embeddings to make them suitable for clustering.

\begin{table}
    \centering
    \textbf{Random Static vs. Dynamic Performance}
    \begin{tabular}{|c|c|c|c|}
        \hline\hline
        Method & MNIST & Fas.-MNIST & CIFAR10 \\ \hline
        % Raw Clustering & 0.0000 & 0.0000 & 0.0000  \\ \hline
        Random Embedding & 0.8473 & 0.8290 & 0.6714 \\\hline
        Semi-Supervised & 0.8700 & 0.8604 & 0.6576\\ \hline
        Dynamic & 0.9113 & 0.8695 & 0.7319 \\
        \hline
    \end{tabular}
    \caption{Comparison of Label Rediscovery Accuracy for various methods.}
    \label{tab:static_vs_dynamic}
\end{table}

\subsection{Application to Language}

We next show that the class discovery method also applies to the language modality. 
In particular, we consider the real-world task of intent detection in goal-oriented dialog modeling, where the goal is to classify a given user utterance to one of the many pre-defined task categories (i.e., intents), so that the dialog manager can direct the downstream flow of the conversation accordingly toward certain task fulfillment modules \citep{zhao_review_2019}. 
In this context, the ability of a dialog agent to automatically discover novel user intent is important for the continuous improvement and refinement of the system. For example, the discovered novel intent types (see, e.g., Table \ref{tb:new_intent}) can be used by the dialog system to implement new response and behavior for the discovered class.

We consider the CLINC Out-of-scope (OOS) benchmark \citep{larson_evaluation_2019}, which contains 150 intent categories with 150 utterance in each category, and an extra 1200 out-of-domain utterances which do not match any of the known intents. For the representation learner,  we consider the standard 1-layer, 128-hidden-unit TextCNN \citep{kim_convolutional_2014} with filter sizes 3, 4, 5 and initialize the word embeddings using GloVe \cite{pennington_glove_2014}. We train the model on 120 randomly-sampled classes from the OOS dataset, and evaluate the quality of the learned representation on the rest of the 30 classes. The model was trained for 200 epochs with minibatch size 128 and step size 0.001 using Adam optimizer, and reached a final classification accuracy of 0.9466. For class discovery, we apply K-means with 30 clusters on the TextCNN's hidden representation for the 30 hold-out classes, and compute the label rediscover accuracy of the resulting clusters against the ground truth labels.

To evaluate the benefit of semi-supervision to  class discovery, we compare the result against five popular or the state-of-the-art unsupervised approaches in text domain. Specifically, we consider  the classic methods of Latent Dirichlet Allocation (LDA) and Non-negative Matrix Factorization (NMF), which are based on the term frequency (TF) and term frequency–inverse document frequency (TF-IDF) representations of the utterances, respectively. Next, we consider two standard token-level embedding approaches word2vec \citep{mikolov_distributed_2013} and GloVe \citep{pennington_glove_2014}, where we compute the sentence-level embedding using the  TF-IDF-weighted average of the token-level embeddings. Finally, we consider the state-of-the-art sentence embedding approach Universal Sentence Encoder (USE) \cite{larson_evaluation_2019}. Universal Sentence Encoder is pre-trained jointly on large-scale web corpuses and on natural language inference tasks, and has illustrated state-of-the-art performance in multiple natural language tasks including semantic textual similarity, opinion polarity classification and phrase level sentiment analysis. For all the  embedding methods, we perform clustering using  K-means with 30 clusters.

\begin{figure}[ht]
\includegraphics[width=.99\columnwidth]{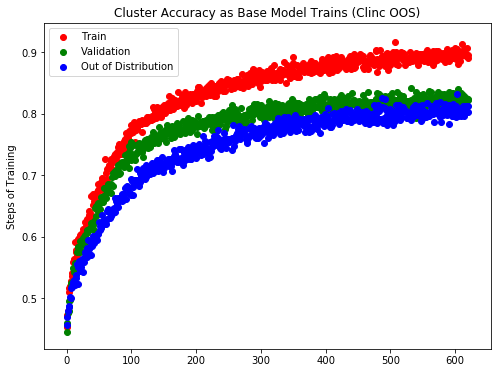}
\caption{The red and green curves measure the correspondence of in-distribution clusters to the original class that has the maximum overlap with the cluster. The blue curve measures the correspondence of newly discovered out-of-distribution clusters to the held-out classes. Measures occur at every 10th step of training.}
\end{figure}

We repeat each of the above methods for 20 times and report their mean and standard deviation of label rediscover accuracy in Table \ref{tb:text_res}. As shown, the class discovery accuracy based on the semi-supervised TextCNN embedding outperforms the both classic and token-level approaches by a clear margin, and is on-par with  Universal Sentence Encoder, despite not being  pre-trained on large-scale corpus like USE.
In particular, we notice that the GloVe-initialized TextCNN outperforms the original GloVe by around $5\%$, illustrating the benefit of semi-supervision in  improving the representation quality of the learned embedding.

\begin{table}[ht]
    \centering
    \begin{tabular}{|c|c|}
        \hline\hline
        Method & Label Rediscovery Accuracy \\ \hline
        Random (Baseline) & 0.2105  \\ \hline
        LDA & 0.5712 $\pm$ 0.0066\\         
        NMF & 0.7478 $\pm$ 0.0066 \\\hline
        word2vec & 0.7194 $\pm$ 0.0089 \\ 
        GloVe & 0.7629 $\pm$ 0.0085 \\ \hline
        Univ. Sent. Encoder & 0.8127 $\pm$ 0.0108 \\ \hline
        TextCNN (Ours) & 0.8123 $\pm$ 0.0076\\
        \hline
    \end{tabular}
    \caption{Comparison with popular unsupervised approaches for class discovery. Mean and Standard Deviation for cluster accuracy on hold-out classes in CLINC intent detection dataset.}
    \label{tb:text_res}
\end{table}

Finally, to illustrate the proposed approach's practical utility in discovering novel intent domains, we perform class discovery on the out-of-scope utterances in the OOS data. Specifically, we apply K-means to the TextCNN representation of the out-of-scope utterances, with the number of clusters determined using the elbow method based on silhouette score \citep{de_amorim_recovering_2015}. We estimate the learnability of the discovered clusters based on the out-of-sample classification accuracy of a logistic regression model, and report the top clusters (with classification accuracy $> 0.95$). We show example utterances in the top clusters in Table \ref{tb:new_intent} and report the full content of the discovered clusters in the Supplementary. We observe that the learnability heuristic is able to select utterance clusters with consistent lexical and syntactical patterns, which tend to correspond to groups of semantically coherent user requests in the context of spoken dialogues. How to incorporate further \textit{a priori} common sense knowledge (e.g., knowledge graph groundings) into this semi-supervised framework to further improve the semantic precision of the discovered utterance  clusters an interesting avenue for future research.

\begin{table}[ht]
    \centering
\begin{adjustbox}{width=0.5\textwidth,center}
    \begin{tabular}{|c|l|}
    \hline\hline
    Intent Domains & Example Utterances \\
    \hline
    \multirow{5}{*}{Anecdote}
       & who is the coach of the chicago bulls \\
       & who formulated the theory of relativity\\
       & can you tell me who sells dixie paper plates\\
       & who invented the internet\\
       & which company has gone up the most today\\\hline
    \multirow{5}{*}{Text Messages}
       & read text\\
       & please read me the last text message i received \\
       & read my friend's text message \\
       & please read the text message i just received\\
       & save my text on my laptop hard drive\\\hline
     \multirow{5}{*}{Phone Inquiries}
    & what's the average battery life of an android phone \\
    & how do i make my android phone more secure \\
     &  how does my current htc phone compare to other android phones \\
     &  what battery percentage is my phone at \\
     &  what are some good games for my android phone \\  \hline
    \multirow{5}{*}{Bank Account}
     & i jot got hired and need help with my retirement account \\
     & i have to roll over my 401k to a new account and i don't know how \\
     & do you know if it is possible to close my savings account \\
     & please take all my money out of my checking account and close the account \\
     & set a warning for when my bank account starts running low \\  \hline
    \hline\hline
    \end{tabular}
    \end{adjustbox}
    \caption{Novel intent domains discovered in the out-of-scope utterances in CLINC intent detection dataset.}
    \label{tb:new_intent}
\end{table}

% \subsection{Ablation}

\section{Conclusion}

Class Discovery is an underserved problem which shows promise to become a major self-supervised learning and semi-supervised learning application. Its success can lead to a new major training style for both language and vision tasks.

We have demonstrated the value of semi-supervised learning to discover a human prior for labeling data. We show that self-supervised learning can continually advance the quality of the representation learned by our model and leveraged that representation to improve the self-supervised process, leveraging a novel heuristic that takes advantage of the clusters that the easiest to learn. Our Dataset Reconstruction Accuracy makes it straightforward to compare different class discovery systems, and we use it to demonstrate our progress on this problem. This is one step forward in what we hope is a impactful future for class discovery.

\bibliography{main}
\bibliographystyle{icml2019}

\end{document}